\documentclass[10pt,twocolumn,letterpaper]{article}

\usepackage{cvpr}              % To produce the CAMERA-READY version
\usepackage{multirow}
\usepackage{tabularx}
\usepackage{comment}
\usepackage{xspace}
\usepackage{float}
\usepackage{bbding}
\usepackage{chapterbib}
\usepackage{placeins}
\usepackage[table,xcdraw]{xcolor}
\newcommand{\printfnsymbol}[1]{%
  \textsuperscript{\@fnsymbol{#1}}%
}

\newcommand{\wu}[1]{\textcolor[rgb]{0,0,0} {#1}}

\newcommand{\lyu}[1]{\textcolor[rgb]{0,0,0} {#1}}

\newcommand{\mymodel}{\textbf{RS-STE}\xspace}

\definecolor{cvprblue}{rgb}{0.21,0.49,0.74}
\usepackage[pagebackref,breaklinks,colorlinks,allcolors=cvprblue]{hyperref}

\def\paperID{15091} % *** Enter the Paper ID here
\def\confName{CVPR}
\def\confYear{2025}

\title{Recognition-Synergistic Scene Text Editing\vspace{-20pt}}

\author{Zhengyao Fang\thanks{Equal contribution.}\textsuperscript{\ \ \ 1}, Pengyuan Lyu\textsuperscript{* 3}, Jingjing Wu\textsuperscript{* 4}, Chengquan Zhang\textsuperscript{3}, Jun Yu\textsuperscript{1},\vspace{2pt}\\
\vspace{4pt}
Guangming Lu\textsuperscript{1}, Wenjie Pei\thanks{Corresponding author.}\textsuperscript{\ \ \ 1, 2}\\
\normalsize{\textsuperscript{1}Harbin Institute of Technology, Shenzhen}\vspace{-2pt}\\
\quad\normalsize{\textsuperscript{2}Peng Cheng Laboratory}\vspace{-2pt}\\
\quad\normalsize{\textsuperscript{3}Tencent}\vspace{-2pt}\\ 
\quad\normalsize{\textsuperscript{4}Department of Computer Vision Technology, Baidu Inc.}\vspace{-2pt}\\
{\tt{\small{\{zhengyaonineve, jingjingwu\_hit, wenjiecoder\}@outlook.com}}}\\
{\tt\small{lvpyuan@gmail.com, zchengquan@gmail.com,}}
{\tt{\small{\{yujun, luguangm\}@hit.edu.cn}}}}
\begin{document}
\maketitle
\begin{abstract}
Scene text editing aims to modify text content within scene images while maintaining style consistency. Traditional methods achieve this by explicitly disentangling style and content from the source image and then fusing the style with the target content, while ensuring content consistency using a pre-trained recognition model. Despite notable progress, these methods suffer from complex pipelines, leading to suboptimal performance in complex scenarios. In this work, we introduce Recognition-Synergistic Scene Text Editing (\mymodel), a novel approach that fully exploits the %intrinsic characteristics of the recognition model. 
intrinsic synergy of text recognition for editing. Our model seamlessly integrates text recognition with text editing within a unified framework, and leverages the recognition model's ability to implicitly disentangle style and content while ensuring content consistency. Specifically, our approach employs a multi-modal parallel decoder based on transformer architecture, which predicts both text content and stylized images in parallel. Additionally, our cyclic self-supervised fine-tuning strategy enables effective training on unpaired real-world data without ground truth, enhancing style and content consistency through a twice-cyclic generation process. Built on a relatively simple architecture, \mymodel achieves state-of-the-art performance on both synthetic and real-world benchmarks, and further demonstrates the effectiveness of leveraging the generated hard cases to boost the performance of downstream recognition tasks. Code is available at \href{https://github.com/ZhengyaoFang/RS-STE}{https://github.com/ZhengyaoFang/RS-STE}.
\end{abstract}    
\section{Introduction}
\label{sec:intro}
\begin{figure}[t]
\centering
    \includegraphics[width=1\linewidth]{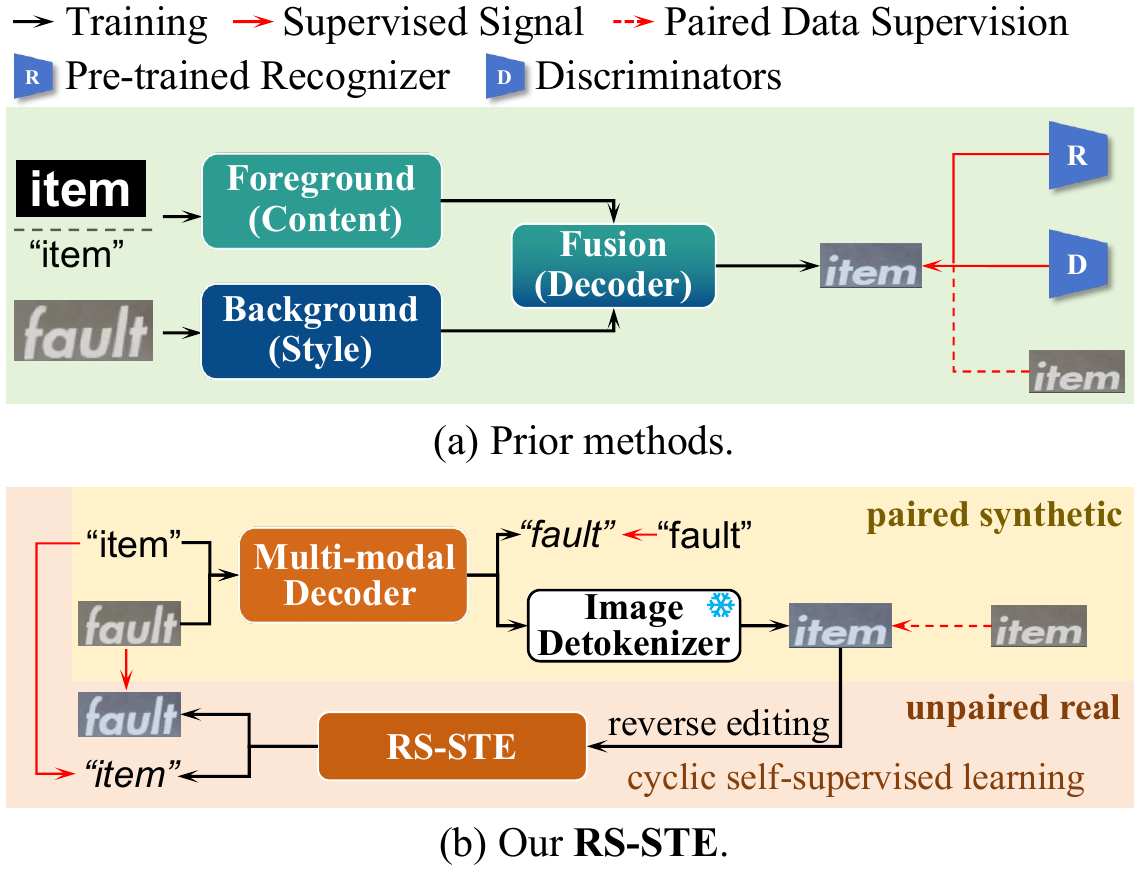}
\caption{Prior methods for scene text editing involve intricate modeling for explicit separation of text content and background style. In contrast, our \mymodel conducts synergistic modeling of scene text recognition and text editing in a unified framework, which allows for implicit text-style separation while ensuring content consistency. Besides, the specially designed Cyclic Self-supervised Fine-tuning enables effective training of \mymodel on unpaired real-world data, substantially enhancing the generalizability in real-world scenarios.}
\label{fig:compair} 

\end{figure}
\lyu{Scene Text Editing (STE) aims to modify the textual content in scene text images while preserving the original style. This technology holds significant potential, enabling designers to efficiently edit and replace textual information in images. Additionally, it can be applied to image generation, thereby enhancing the performance of other Optical Character Recognition (OCR) tasks, such as text detection and recognition. Given its importance, STE has garnered increasing attention from researchers.}

\lyu{STE presents two primary challenges. Firstly, the diverse appearances of scene text, including variations in background, font, and layout, pose significant difficulties for STE. Secondly, the lack of paired real training data necessitates training existing methods on synthetic data. The domain gap between synthetic and real data hinders the ability of models trained on synthetic data to generalize effectively to real-world scenarios.}
\begin{figure}
\centering
    \includegraphics[width=1\linewidth]{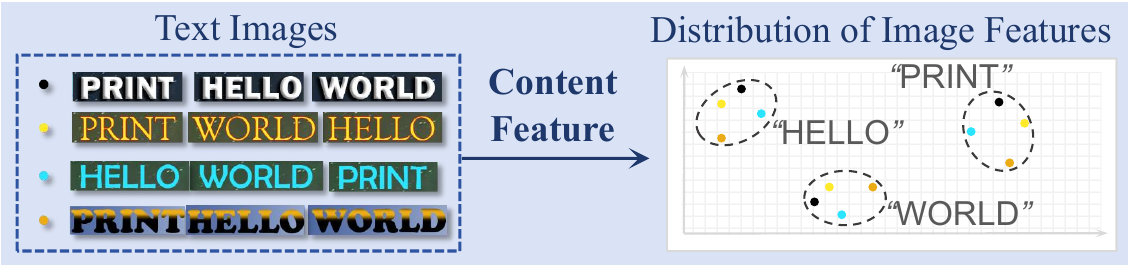}
\caption{Distribution of some content features extracted by our \mymodel. Images with the same text content but different background styles become closer in the encoded feature space of a recognition model, implying the capability of recognition models to separate style from content.}
\label{fig:recognition} 

\end{figure}
\lyu{Many methods have been proposed to address the aforementioned challenges. Most of these methods follow a pipeline that involves explicitly decomposing the information in the source image into style and content components, and then merging the source style with the target text content to produce the target image. A pre-trained recognition model is always used to ensure the content consistency in the edited image. For instance, as illustrated in Figure~\ref{fig:compair} (a), existing approaches~\cite{wu2019editing,roy2020stefann,yang2020swaptext,qu2023exploring,DBLP:conf/iclr/DongHPQGYZSZWK024} typically involve an explicit separation of the background and foreground images. Other methods~\cite{lee2021rewritenet, zhang2024choose} focus on learning to disentangle the content features and style features from the source style image. Furthermore, to address the issue of lacking real paired data, several learning mechanisms~\cite{qu2023exploring,lee2021rewritenet, yang2023self,krishnan2023textstylebrush} have been proposed. These mechanisms employ a combination of real and synthetic data during training to enhance the model's generalization capabilities in real-world scenarios. By leveraging both types of data, these approaches aim to bridge the domain gap and improve the performance of scene text editing models when applied to real-world images. Most of these methods, when training on unpaired real data, follow a paradigm similar to the illustration in Figure~\ref{fig:compair} (a), where paired data supervision is not available.}

\lyu{While previous methods have made remarkable progress, their intricate pipelines introduce two potential limitations that hinder further performance improvements. Firstly, explicitly separating style and content is a challenging task and may not always be perfectly accurate, which can result in suboptimal outcomes when these components are recombined. Secondly, these methods often consist of multiple interconnected modules, and the joint optimization of these modules can also lead to less-than-ideal results.}

\lyu{In this paper, we introduce \mymodel{}, a novel approach that not only addresses the limitations of existing methods but also delivers superior editing outcomes. The impetus for our approach arises from a fundamental observation: recognition models inherently separate style from content, as illustrated in Figure~\ref{fig:recognition}. By capitalizing on this characteristic, we seamlessly integrate the recognition model with the editing model within a cohesive framework. Specifically, we have developed a multi-modal parallel decoder based on the transformer decoder architecture. This decoder, upon receiving the encoded tokens of the specified text and the source style image, concurrently predicts the text content on the style image and generates an image with the source style and specified text. Furthermore, we propose a Cyclic Self-Supervised Fine-tuning mechanism, as illustrated in Figure~\ref{fig:compair} (b) for \textit{unpaired real} data, to effectively harness real-world data for training purposes. This design maximizes the potential of the recognition model in two principal ways: (1) It implicitly decouples style and content, allowing the model to better capture attributes for generating target image, and (2) Within the Cyclic Self-Supervised Fine-tuning, the recognition model's supervision ensures the consistency of the generated content. As a result, our approach offers a significant advantage: it eliminates the need for multiple modules to explicitly decouple style and content, and it obviates the necessity for a separate recognition model to verify the accuracy of the generated content. This greatly simplifies the overall pipeline and circumvents the challenges faced by previous methodologies.}

To conclude, our contributions are listed as follows:
\begin{itemize}[leftmargin =*, itemsep = 0pt, topsep = -2pt]
    \item We propose a simple yet effective scene text editing method dubbed \mymodel, which conducts recognition-synergistic scene text editing in a unified framework. Such design enables implicit separation between background style and text content, thereby eliminating intricate model design.%\wu{We propose a novel scene text editing pipeline \mymodel{} based on transformer decoder (rather than separating style and content).}
    \item We design the Cyclic Self-supervised Fine-tuning Strategy, which allows for effective training on unpaired real-world data to substantially enhance its generalizability in real-world scenarios. %apply supervision to edited images of real-world data to  ensure style  and content consistency.%, resulting state-of-the-art performance on real-world benchmark.}
    \item
    % In order to better evaluate style consistency on real-world data, we proposed the gpt-4o evaluation protocol. 
    Our \mymodel achieves state-of-the-art performance on both synthetic and real-world scene text editing benchmarks. We further validated the effectiveness of our generated image on downstream recognition tasks.%\wu{recognition benefits editing task (unified transformer framework instead of a stand-alone recognition model).}
    
\end{itemize}

\section{Related Work}
\label{sec:related_work}
\subsection{Scene Text Editing}
\wu{Scene Text Editing refers to the process of modifying text within natural scene images while maintaining the visual consistency and context of the surrounding elements. Early work designed complex modules to explicitly separate foreground and background. SRNet~\cite{wu2019editing} leverages three respective modules for learning background reconstruction, render text, and final fusion. Similarly, STEFANN~\cite{roy2020stefann} focuses on character-wise rendering through a text conversion and color transfer module, applying them to the inpainted background. Advancing these foundations, SwapText~\cite{yang2020swaptext} introduces a spatial transformation to adapt to oriented text. To implicitly decouple the content and style features and use the unpaired real-world data for training, TextStyleBrush~\cite{krishnan2023textstylebrush} propose to use task-adaptive StyleGAN2~\cite{karras2020analyzing} along with self-supervised training strategy. RewriteNet~\cite{lee2021rewritenet} learns to separate content and style features and fine-tunes on real-world images with a self-supervised training scheme. MOSTEL~\cite{qu2023exploring} intentionally focuses on style by discarding content information, using style augmentation techniques to merge style and content images, thus producing target content with a reconstructed background. CLASTE~\cite{yang2023self} uses extra background restoration module for background restoration and integrates the background with foreground content. DARLING~\cite{zhang2024choose} combines aligned style and content features using a multi-task decoder enhanced by self-attention blocks, showcasing a blend of synthesis, self-supervised learning, and contextual awareness in modern scene text editing techniques. Recently, STEEM~\cite{DBLP:conf/iclr/DongHPQGYZSZWK024} introduces a minimized background reconstruction technique to explicitly decouple the style and content and further enhance text editing fidelity.}

\wu{Recently, several works ~\cite{ji2023improving,chen2024diffute,chen2024textdiffuser,tuo2023anytext} explore the manipulation of text within the entire image using stable diffusion~\cite{rombach2022high}. They encompass the intricate aspects of layout arrangement and the accurate rendering of textual content. As this focus differs from the concerns addressed in our paper, we have not conducted a comparison with our work.}
%Additionally, a task closely related to scene text editing is universal text editing, which enables modifications to any text across an entire image containing multiple text elements, rather than requiring cropped text images. Most existing studies in this area\cite{ji2023improving, chen2024diffute, chen2024textdiffuser, tuo2023anytext} rely on diffusion-based methods and is not directly comparable to the approach investigated in this study. 

\subsection{MLLM for Image Generation and Editing}
In response to the notable progress of large language models in natural language processing~\cite{achiam2023gpt,chowdhery2023palm, bai2023qwen}, the field of multi-modal large language models (MLLM) has made significant strides in recent years. MLLMs leverage both natural language and visual inputs, allowing these models to understand and manipulate visual data guided by textual descriptions. This dual-modality capability builds upon foundational image generation models, such as GANs~\cite{goodfellow2014generative} and diffusion models~\cite{ho2020denoising}, but advances them by incorporating language as a critical component in model design. Recent works~\cite{ge2024seed, team2023gemini, sun2024generative, li2024mini, achiam2023gpt} have developed architectures capable of processing text and image modalities simultaneously, achieving a more nuanced integration of linguistic and visual information. These approaches demonstrate enhanced performance in image generation tasks, where MLLMs generate high-quality visuals that align closely with the semantic content of textual prompts. Furthermore, some MLLMs~\cite{hertz2022prompt, mokady2023null, brooks2023instructpix2pix} offer innovative capabilities for image editing by enabling users to adjust existing images through descriptive language, such as modifying attributes or inserting new elements, rather than relying on pixel-level manipulation. Inspired by these methods, our approach integrates the multi-modal language model \mymodel, which is specialized in scene text editing.

\section{Method}
\begin{figure}[t]
\centering
    \includegraphics[width=1\linewidth]{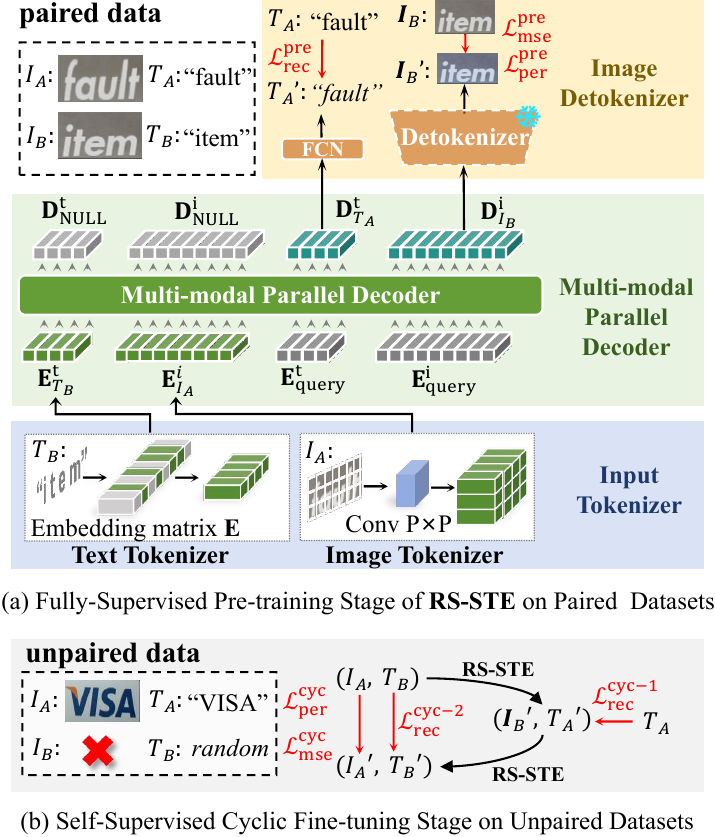}
\caption{(a) illustrates the model structure of \mymodel and the fully-supervised pre-training stage using paired synthetic datasets. (b) depicts the cyclic self-supervised fine-tuning stage with unpaired real-world datasets.}
\label{fig:model} 

\end{figure}

\label{sec:method}
\subsection{Overview}
The aim of scene text editing is to edit text image $I_A$ to synthesize image $I_B$ by altering the text content $T_A$ into the desired content $T_B$ while retaining the image style of $I_A$. Our proposed \mymodel for this task is able to conduct text recognition and editing within a unified framework, resulting in a straightforward pipeline. As shown in Figure~\ref{fig:model}, it comprises Input Tokenizer, Multi-modal Parallel Decoder, and Image Detokenizer.

Given the target text content $T_B$ and a reference image $I_A$, Input Tokenizer encodes them into text embeddings and image embeddings respectively, and outputs a cascaded embedding sequence. Then Multi-modal Parallel Decoder performs scene text editing in the feature space and predicts the tokens of $T'_A$ and $I'_B$ in a parallel manner. Lastly, Image Detokenizer generates target image $I'_B$ from decoded image tokens $\textbf{D}^{\text{i}}_{I_B}$. While the generated $I'_B$ contains different text content from $I_A$, their stylistic components including background and typeface are required to be completely identical. 

Our \mymodel is optimized in two learning stages. It is first trained on a large corpus of synthesized data with paired $I_A$ and $I_B$ to endow it with the basic capability of scene text editing. Then in the second stage, it is further optimized on unpaired real-world data (without ground-truth) using our specially designed cyclic self-supervised fine-tuning strategy, which substantially improves its robustness and generalizability towards real-world data. We will first describe the model structure of \mymodel in Section~\ref{sec:model_pipeline}, and then elaborate on the training strategy in Section~\ref{sec:stage2}.  

%we can acquire the capability of manipulating of text images in accordance with specified text content and a predetermined style. In addition, to effectively utilize more unpaired real-world data for training, we further propose Cyclic Self-supervised Fine-tuning strategy. In this section, we will first describe the network structure of \mymodel{} in Section~\ref{sec:model_pipeline}, and then elaborate on the training strategy in Section~\ref{sec:stage2}.

% Given the style image I_S, and the target text content T_T, our input tokenizer will tokenize I_S and T_T respectively, and input them into the Parallel Decoder. The Parallel Decoder predicts the text content T_S of the style image I_S and the image token sequence I_T containing the text content T based on the input image and text information at the output end.
%Our \mymodel{} is constructed on the foundation of a multimodal transformer decoder and employs a multi-stage training strategy comprising three key phases: (1) fine-tuning the decoder on our training datasets using a pre-trained autoencoder\footnote{https://github.com/CompVis/latent-diffusion}, as outlined by \cite{esser2021taming};  (2) conducting pretraining of \mymodel{} on paired synthesis datasets in a fully supervised manner; and (3) refining the model on real datasets in a self-supervised manner through circular editing, inspired by the concept in \cite{zhu2017unpaired}. We first describe the structure of \mymodel{} in Section \ref{sec:model_pipeline}, before concentrating on the training stages detailed in Sections \ref{sec:stage2} and \ref{sec:stage3}.
\subsection{\mymodel}
\label{sec:model_pipeline}
\smallskip\noindent\textbf{Input Tokenizer.}  The Input Tokenizer encodes the input target text $T_B$ and the reference style image $I_A$ separately. For text encoding, we learn an embedding matrix $\mathbf{E}\in \mathbb{R}^{(|\Sigma| + 1) \times C}$ for alphabet $\Sigma$, from which we can encode $T_B = \{c_1, \dots, c_L\}$ by indexing the corresponding character embeddings sequentially, thereby obtaining the text embedding $\textbf{E}^{\text{t}}_{T_B} \in \mathbb{R} ^{L \times C}$. 

We adopt the ViT-based tokenization approach to encode the reference style image $I_A \in \mathbb{R}^{H \times W \times 3}$. To be specific, we apply a convolutional layer with a kernel size of $P \times P$ to split image into $\frac{H}{P} \times \frac{W}{P}$ patches and capture visual information, producing flattened visual feature sequence $\textbf{E}^{\text{i}}_{I_A} \in \mathbb{R}^{N \times C}$, where $N = (HW) / P^2$.

%$X_A \in \mathbb{R}^{{\frac{H}{P}} \times {\frac{W}{P}}  \times C}$. }
%where $H$, $W$, $C$ refer image height, image width and channels respectively
% we use a (4, 4) convolution layer to obtain [H // 4, W //4, C], and further reshape it into (N, C), where N is the length of the image feature sequence, equal to H // 4 * W //4.

% the overview of Parallel Decoder
\smallskip\noindent\textbf{Multi-modal Parallel Decoder (MMPD).} A scene text recognition model is capable of extracting text-related features from an image by implicitly distinguishing between text and background style. In light of this, instead of disentangling style and text content via separate task modeling as other methods~\cite{wu2019editing,roy2020stefann,yang2020swaptext,qu2023exploring,DBLP:conf/iclr/DongHPQGYZSZWK024} perform, our \textbf{RS-STE} model conducts both scene text recognition and text editing in the unified Multi-modal Parallel Decoder to leverage the synergy of text recognition towards editing. As shown in Figure~\ref{fig:model}, given $\textbf{E}^{\text{t}}_{T_B}$ and $\textbf{E}^{\text{i}}_{I_A}$, the Multi-modal Parallel Decoder is optimized to recognize the text content $T_A'$ while performing text editing in the feature space to predict the token features of the target image $I_B'$.

The Multi-modal Parallel Decoder is designed in the structure of Transformer decoder. Following the classical modeling paradigm of multi-modal language foundation models~\cite{dong2023dreamllm,wu2023next,ge2024seed}, we initialize learnable query embeddings corresponding to the text and image prediction, denoted as $\mathbf{E}^{\text{t}}_{\text{query}} \in \mathbb{R}^{L \times C}$ and $\mathbf{E}^{\text{i}}_{\text{query}} \in \mathbb{R}^{N \times C}$ respectively. They are sequentially concatenated after $\mathbf{E}^\text{t}_{T_B}$ and $\mathbf{E}^\text{i}_{I_A}$, and fed into the Multi-modal Parallel Decoder:
\begin{equation}
\begin{split}
&[\mathbf{D}^{\text{t}}_\text{NULL}, \mathbf{D}^{\text{i}}_\text{NULL}, \mathbf{D}^\text{t}_{T_A}, \mathbf{D}^\text{i}_{I_B}] \\
= &\mathcal{F}_\text{MMPD}([\mathbf{E}^\text{t}_{T_B}, \mathbf{E}^\text{i}_{I_A}, \mathbf{E}^\text{t}_\text{query}, \mathbf{E}^\text{i}_\text{query}]).
\end{split}
\end{equation}
where $\mathbf{D}^\text{t}_{T_A} \in \mathbb{R}^{L\times C}$ and $\mathbf{D}^\text{i}_{I_B} \in \mathbb{R}^{N\times C}$ are the decoded token features for $T_A$ and $I_B$, respectively. It is noteworthy that the first $(L + N)$ output tokens aligned with $\mathbf{E}^\text{t}_{T_B}$ and $\mathbf{E}^\text{i}_{I_A}$, denoted as $\mathbf{D}^{\text{t}}_\text{NULL}$ and $\mathbf{D}^{\text{i}}_\text{NULL}$, are not used since they cannot access the full content of $\mathbf{E}^\text{t}_{T_B}$ and $\mathbf{E}^\text{i}_{I_A}$. $\mathbf{D}^\text{t}_{T_A}$ is further used to perform text recognition by a fully connected layer (FCN) that predicts the character probabilities $\mathbf{P}_{T_A}\in \mathbb{R}^{L\times {(|\Sigma| + 1})}$, while $\mathbf{D}^\text{i}_{I_B}$ is fed into the Image Detokenizer to synthesize the edited image $I_B$.

\smallskip\noindent\textbf{Image Detokenizer.}  We utilize the pre-trained VAE decoder of LDM~\cite{rombach2022high} as the Image Detokenizer and fine-tune it on the synthesized training data. Following the routine training paradigm~\cite{esser2021taming}, the Image Detokenizer is fine-tuned before the training of the Input Tokenizer and Multi-modal Parallel Decoder of \mymodel for stable optimization. 
\begin{figure*}
\centering
    \includegraphics[width=0.7\linewidth]{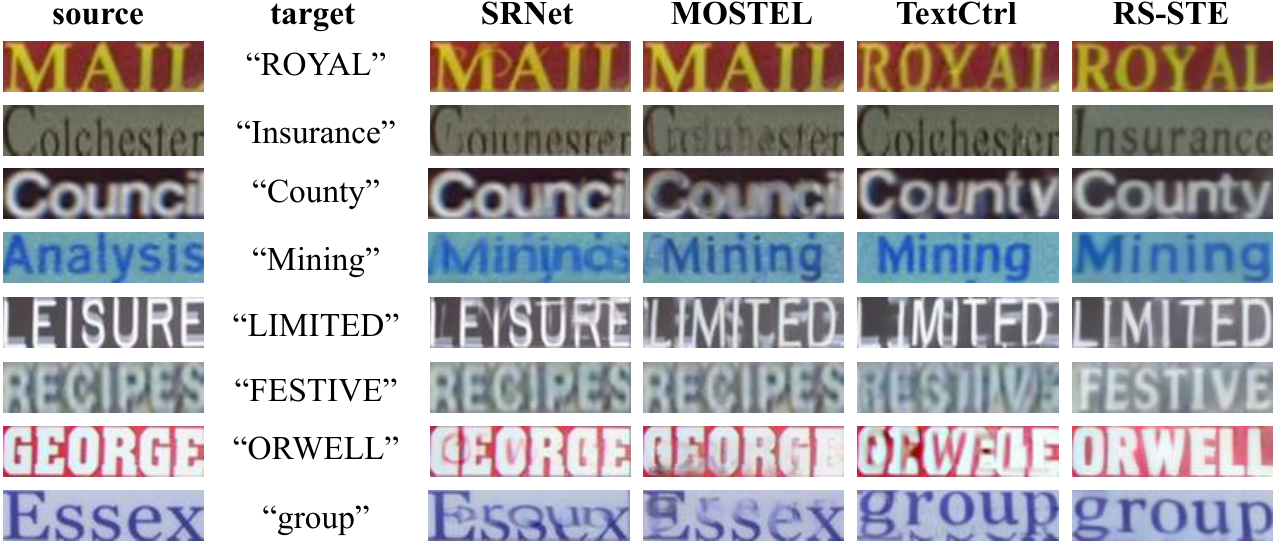}
\caption{ Editing examples compared with other methods.}
\label{fig:examples} 
\vspace{-15pt}
\end{figure*}
\subsection{Training Strategy}
\label{sec:training}
Our \mymodel is optimized in two stages: a fully-supervised pre-training stage on paired synthetic data and a cyclic self-supervised fine-tuning stage on unpaired real-world data.

\smallskip\noindent\textbf{Fully-Supervised Pre-training Stage.} 
\label{sec:stage2}
As collecting paired real-world data for supervised scene text editing is infeasible, we first pre-train our \mymodel on synthetic paired data to equip it with the fundamental capability of scene text editing. Since our model is able to perform synergistic modeling of both scene text recognition and text editing, we conduct supervised learning on both tasks, as illustrated in Figure~\ref{fig:model} (a). 
To be specific, we adopt cross-entropy loss to optimize scene text recognition:

\begin{equation}
    \mathcal{L}_{\text{rec}}(T_A, T_A') =  -\frac{1}{L} \sum_{i=1}^{L} \sum_{c=1}^{|\Sigma|+1} G(T_A)[i, c] \log(\mathbf{P}_{T_A}[i, c]),
    \label{eqn:rec}
\end{equation}
where $L$ is the max length of text. $G(T_A)[i, c]$ is the one-hot encoding ground truth at position $i$, with  the $c$-th character in the pre-defined alphabet equal to 1, while $\mathbf{P}_{T_A}[i, c]$ is the corresponding prediction by our \mymodel.

To supervise scene text editing, we employ the mean squared error (MSE) loss for pixel-level supervision and perceptual loss~\cite{johnson2016perceptual} for semantic alignment between the edited image and the ground truth. Formally, for the edited image $I_B'$, the MSE loss and perceptual loss are defined as:
\begin{equation}
\begin{split}
    &\mathcal{L}_{\text{mse}}(I_B, I_B') = \Vert I_B-I_B'\Vert_2^2,\\
    &\mathcal{L}_{\text{per}}(I_B, I_B') = \mathbb{E} \left [\Vert \phi_i(I_B)-\phi_i(I_B') \Vert_2^2 \right ],
\end{split}
    \label{eqn:mse}
\end{equation}
where $\phi_i$ is the features extracted from \textit{relu1\_2}, \textit{relu2\_2}, \textit{relu3\_3} and \textit{relu4\_3} of a pre-trained VGG-16 network~\cite{simonyan2014very}. 

Integrating all three losses, the overall learning objective for our \mymodel in the pre-training stage is formulated as:
\begin{equation}
\begin{aligned}
    \mathcal{L}^\text{pre} = \lambda_1\mathcal{L}^\text{pre}_\text{\text{rec}}(T_A, T_A') 
 +  \lambda_2\mathcal{L}^\text{pre}_\text{\text{mse}}(I_B, I_B')  + \lambda_3\mathcal{L}^\text{pre}_\text{\text{per}}(I_B, I_B'),\\
\end{aligned}%
\end{equation}
where $\lambda_1$, $\lambda_2$ and $\lambda_3$ are balancing weights, and the superscript `pre' indicates that these losses only apply to the pre-training stage.

\smallskip\noindent\textbf{Cyclic Self-Supervised Fine-tuning Stage.}
\label{sec:stage3}
Despite the abundance of paired synthetic training data available for pre-training, the significant disparity between synthetic data and real-world data severely limits the applicability of the pre-trained model in real-world scenarios. Nevertheless, conducting direct supervised learning with real-world data is impractical in the absence of paired data for scene text editing. To circumvent this problem, we devise the Cyclic Self-Supervised Fine-tuning strategy, which conducts scene text editing twice on a same style image with reverse operation by interchanging the target text, reproducing the initial style image. As shown in Figure~\ref{fig:model} (b), given a style image $I_A$ and the target text $T_B$, our \mymodel generates $I_B'$ and predicts $T_A'$ in the first scene text editing. Then, using $I_B'$ and $T_A'$ as the style image and target text respectively, \mymodel performs the second editing and obtains the edited image $I_A'$ and recognized text$T_A'$, which should be the reproduction of the initial style image $I_A$. The whole process can be expressed as:
\begin{equation}
    \begin{split}
    &(I_B',T_A')=\mathcal{F}_\text{\mymodel}(I_A,T_B),\\
    &(I_A', T_B')=\mathcal{F}_\text{\mymodel}(I_B',T_A'),
    \end{split}
\end{equation}
where $\mathcal{F}_\text{\mymodel}$ denotes editing function by our \mymodel. The proposed cyclic editing procedure allows us to perform supervision on the reproduced image $I_A'$, which is equivalent to self-supervised learning.

\begin{comment}
\wu{We propose Cyclic Self-Supervised Fine-tuning strategy that performs pixel-wise supervision on the generated image in an indirect manner, enabling \mymodel{} to successfully handle complex styles in real-world data. As illustrated in Figure\ref{fig:model} (b), given the absence of paired $I_A$ and $I_B$ in real data, and the availability of only a large quantity of $I_A$ and $T_A$, we initially generate $I_B'$ by randomly selecting $T_B$ that of similar length of $T_A$. Once we have obtained generated $I_B'$, we can proceed to generate $I_A'$ and $T_B'$ using \mymodel. By supervising the two generation processes of \mymodel, we can circumvent the problem of lacking $I_B$.}

\wu{Formally, we use the $\mathcal{F}_\text{\mymodel}$ denote \mymodel{} and the first generation process can be represented as: }

\begin{equation}
    (I_B',T_A')=\mathcal{F}_\text{\mymodel}(I_A,T_B).
\end{equation}

\wu{The predicted $T_A'$ can be supervised by recognition loss in Equation ~\ref{eqn:rec} to ensure \mymodel{} can implicitly separate style and text content in $I_A$. In addition, we can consider $I_B'$ has different text content but the same style as $I_A$. During the second generation process, \mymodel{} takes $I_B'$ and $T_A$ as input and finally output $I_A'$ and $T_B'$:}

\begin{equation}
    (I_A', T_B')=\mathcal{F}_\text{\mymodel}(I_B',T_A),
\end{equation}
\end{comment}

In this stage, we also use MSE loss and perceptual loss to supervise the generation of $I_A'$. Meanwhile, we apply the recognition loss to both predicted $T_A'$  and $T_B'$ in twice text editing, which can prevent the model from collapsing into an identical mapping before and after cyclic editing:
%from  employed on the predicted $T_B'$ in order to prevent the model from directly learning a mapping from the original image to itself by retaining all the information of $I_A$ in $I_B'$ directly, we employ an additional . Then the total loss of \underline{C}yclic \underline{S}elf-supervised \underline{F}ine-tuning Stage is formalized as follows:
\begin{equation}
\begin{split}
    \mathcal{L}^\text{cyc} &= \lambda_4\mathcal{L}^\text{cyc}_{\text{mse}}(I_A, I_A')+\lambda_5\mathcal{L}^\text{cyc}_{\text{per}}(I_A, I_A')\\
    &+\lambda_6\mathcal{L}^\text{cyc-1}_{\text{rec}}(T_A, T_A') + \lambda_7\mathcal{L}^\text{cyc-2}_{\text{rec}}(T_B, T_B'),
\end{split}
\end{equation}
where $\lambda_4$, $\lambda_5$, $\lambda_6$ and $\lambda_7$ are hyper-parameters for balancing between different losses. %Note that $\mathcal{L}^\text{pre}_{\text{rec}}(T_A, T_A')$ is used 

\section{Experiment}
\begin{table*}[t]
\caption{Comparison on editing performance with state-of-the-art methods on paired synthetic dataset Tamper-Syn2k, unpaired real dataset Tamper-Scene and paired real dataset ScenePair. The SSIM and SeqAcc are presented in percent (\%).}
\label{tab:pair_editing}
\centering
\resizebox{0.9\linewidth}{!}{%
\begin{tabular}{l|cccc|c|ccccc|c}
\toprule
\multirow{2}{*}{Methods} & \multicolumn{4}{c|}{Tamper-Syn2k}   & Tamper-Scene& \multicolumn{5}{c|}{ScenePair} &\multirow{2}{*}{\#Params}\\ \cmidrule{2-11}
                         & MSE$\downarrow$ & PSNR$\uparrow$ & SSIM$\uparrow$ & FID$\downarrow$ & RecAcc$\uparrow$& MSE$\downarrow$& PSNR$\uparrow$& SSIM$\uparrow$& FID$\downarrow$&RecAcc $\uparrow$  &                \\ \midrule
                    pix2pix     &  0.1450   & 9.18     &  34.15    & 127.21     &  13.26 &$-$ &$-$ & $-$ &$-$ &      $-$     &  54.4        \\
                    SRNet\cite{wu2019editing}     & 0.0216    &   18.66   &  49.97    & 64.37     &  30.26 &0.0561&14.08&26.66&49.22& 17.84 & \textbf{18.95}\\
                    SwapText\cite{yang2020swaptext}     & 0.0194    & 19.43     &  52.43    & $-$      &    54.83&$-$ & $-$&$-$  &$-$ & $-$ &$-$\\
                    MOSTEL\cite{qu2023exploring}     & 0.0135    &  20.27    &  56.94    & 33.79     &  66.54&0.0519&14.46&27.45&49.19&37.69   &54.0  \\
                    % DiffSTE\cite{ji2023improving} & $-$    & $-$ & $-$  & $-$   &  $-$ & 0.0611&13.44&26.85&120.34&31.35\\
                    DARLING\cite{zhang2024choose}     & 0.0120    & 20.80     & 60.07     & 44.48     &  70.85 &$-$ &$-$ &$-$ &$-$ &  $-$ & 23.7 \\ 
                    STEEM \cite{10719657}     & 0.0122    & 20.83     & 72.10     & \textbf{24.67}     &  78.80 &$-$ &$-$ &$-$ &$-$ &  $-$& 69.6\\
                    % TextDiffuser\cite{chen2024textdiffuser} &  $-$   & $-$ & $-$  & $-$   &  $-$&0.0575&13.96&27.02&57.01&51.48\\
                    % AnyText\cite{tuo2023anytext} &   $-$  & $-$ &  $-$ & $-$   &  $-$&0.0619&13.66&30.73&51.79&51.12\\
                    TextCtrl\cite{zeng2024textctrl} &   0.0130  & 20.79 & 66.60  & 31.13   &  74.17&0.0447&14.99&37.56&43.78&84.67&1216.0\\
                    \midrule
                    Ours     & \textbf{0.0076}    &  \textbf{22.54}    &  \textbf{72.90}    & 30.29     &  \textbf{86.12}  &\textbf{0.0267} &\textbf{17.35}&\textbf{46.09}&\textbf{41.37}   &\textbf{91.80}       &  54.4 \\ \bottomrule
\end{tabular}%
}
\vspace{-10pt}
\end{table*}

\begin{table}
\caption{\textbf{Text Recognition Accuracy} on the STR common benchmark datasets. \textit{Base} is provided as baseline results of the recognition model \cite{fang2021read} on the original images, implying the upper bound of recognition performance. Others are the recognition results on the edited datasets generated by different models.}
\label{tab:real_editing}
\resizebox{1\linewidth}{!}{%
\begin{tabular}{l|l|cccccc|c}
\toprule
\multirow{2}{*}{Methods} & Real Training & \multicolumn{6}{c|}{Recognition Benchmarks Accuracy} & \multirow{2}{*}{Avg} \\ \cmidrule{3-8}
                        &Dataset & IIIT   & IC13   & SVT   & IC15   & SVTP   & CUTE   &                      \\ \midrule
Base                    & $-$ &   \textit{96.5}       &   \textit{95.1}     &   \textit{94.7}    &   \textit{85.9}     & \textit{89.5}       &    \textit{89.6}    &       \textit{91.8}               \\ \midrule
TextCtrl          &   $-$     &     70.0     &    68.8    &  73.7     &   62.6     &   63.4    &     58.7   &   66.2                  \\
MOSTEL          &   MLT2017     &    48.2      &    40.6    & 41.1      &       34.4 &    21.6    &  35.1      &      36.8                \\  \midrule
Ours (w/o $\mathcal{L}^{\text{cyc}}$)                & $-$ &   61.2       &    61.2    &   66.8    &    53.8    &   46.4     &    44.8     &          55.7            \\
Ours                   & MLT2017 &   76.6       &    74.8    &   89.6    &    83.0    &   86.4     &    \textbf{80.6}     &          81.8            \\
Ours                & Union14M-L &    \textbf{78.4}      &    \textbf{76.2}    &   \textbf{91.3}    &  \textbf{85.8}      &   \textbf{88.2}     &     77.4   &    \textbf{82.9}                  \\ \bottomrule
\end{tabular}%
}

\end{table}

\subsection{Datasets}
\label{sec:datasets}
\noindent\textbf{Training.} In the pre-training stage, we utilize the open-source synthetic Tamper-train~\cite{qu2023exploring} dataset, which consists of 150k images. Additionally, following~\cite{zhang2024choose}, we incorporate 4M paired synthetic data samples generated using the same image synthesis engine\footnote{https://github.com/youdao-ai/SRNet-Datagen} employed in Tamper-train. During the fine-tuning stage, we utilize the MLT-2017~\cite{nayef2017icdar2017} real dataset in accordance with the MOSTEL~\cite{qu2023exploring} to ensure a fair comparison. \lyu{In order to further explore the effectiveness of our method, we also conducted training on the Union14M-L~\cite{Jiang2023RevisitingST} dataset to verify whether more complex and diverse training data can yield better results.}

% To further explore the validity of our method, we also utilize Union14M-L~\cite{Jiang2023RevisitingST} to obtain better editing results.

\noindent\textbf{Evaluation.}
For evaluation, we use the paired synthetic dataset Tamper-Syn2k~\cite{qu2023exploring} and the paired real-world dataset ScenePair~\cite{zeng2024textctrl} to assess the discrepancies between our edited images and their target counterparts. These two datasets contain 2,000 pairs and 1,280 pairs of text images, respectively, with different text content but the same stylistic components. Additionally, we leverage the unpaired real-world dataset Tamper-Scene~\cite{qu2023exploring}, comprising 7,725 unpaired images, to indirectly evaluate editing quality. Furthermore, we incorporate six commonly used text recognition benchmark datasets—IIIT 5K-Words (IIIT)~\cite{mishra2012scene}, ICDAR2013 (IC13)~\cite{karatzas2013icdar}, Street View Text (SVT)~\cite{wang2011end}, ICDAR2015 (IC15)~\cite{karatzas2015icdar}, Street View Text-Perspective (SVTP)~\cite{phan2013recognizing}, and CUTE80 (CUTE)~\cite{risnumawan2014robust}——for further indirect evaluation of editing quality in more complex and diverse real-world scenarios. Finally, we utilize the Union-benchmark~\cite{Jiang2023RevisitingST} dataset to assess improvements in recognition models facilitated by our targeted data augmentation strategy as detailed in Section~\ref{sec:finetune_app}.

\subsection{Implementation Details}
We initialize the pre-trained VAE~\cite{Kingma2013AutoEncodingVB} with configuration parameters $f=4$, $Z=8192$ and $d=3$ from LDM~\footnote{https://github.com/CompVis/latent-diffusion}. This model is fine-tuned on the Tamper-train dataset, and its decoder is frozen for use in subsequent stages. The minGPT model~\footnote{https://github.com/karpathy/minGPT}, with 22.5M parameters, serves as the foundation of MMPD and is pre-trained from scratch. We adopt AdamW~\cite{loshchilov2017decoupled} as the optimizer with $\beta_1=0.9$ and $\beta_2=0.95$. Following~\cite{zhang2024choose}, the image size in both training and evaluation is set to $32 \times 128$.

In the fully-supervised pre-training stage, we set a batch size of 32 and a learning rate of $1.44\times10^{-4}$, training for a total of 300k iterations. $\lambda_1$, $\lambda_2$, and $\lambda_3$ are set to 1, 10, and 1, respectively. In the subsequent cyclic self-supervised fine-tuning stage, we use a batch size of 16 and a learning rate of $7.2 \times 10^{-5}$, training for 1k iterations. The weights $\lambda_4$, $\lambda_5$, $\lambda_6$ and $\lambda_7$ are set to 10, 1, 50, 50, respectively. All experiments are conducted on 4 NVIDIA 3090 GPUs.
\subsection{Evaluation Metrics}
\label{sec:eval_metrics}
Regarding the paired evaluation datasets Tamper-Syn2k and ScenePair, where the target edited images are specified, we utilize four metrics to assess the differences between the predicted and target images. These metrics include 1) Mean Squared Error (MSE), which measures the $L2$ distance; 2) Peak Signal-to-Noise Ratio (PSNR), representing the ratio of peak signal power to noise power; 3) Structural Similarity Index Measure (SSIM), which evaluates mean structural similarity; and 4) Fréchet Inception Distance~\cite{heusel2017gans} (FID), which calculates the distance between features extracted using the InceptionV3~\cite{szegedy2016rethinking} model. Higher PSNR, SSIM, and lower MSE, FID indicate better performance. 

Moreover, for the real datasets Tamper-Scene and ScenePair, we use the same pre-trained recognition model CRNN~\cite{shi2016end}, as in MOSTEL~\cite{tuo2023anytext} to compute the Recognition Accuracy (RecAcc) of the generated images, which serves as an indirect indicator of the consistency between the text content of the edited images and their target texts.

\subsection{Comparison on Editing Performance}
We perform two sets of experiments to evaluate our method: 1) evaluation on the standard benchmarks for scene text editing including paired synthetic dataset Tamper-Syn2k, paired real dataset ScenePair, and unpaired real dataset Tamper-Scene; 2) experiments on more challenging real datasets, i.e., the classical datasets for text recognition.

Table~\ref{tab:pair_editing} presents the scene text editing performance of different methods in the first set of experiments. Our method achieves the best performance across all evaluation metrics, except for the FID on synthetic data. Specifically, on the paired real dataset ScenePair, \mymodel shows significant improvements in MSE, PSNR, SSIM, and RecAcc. On the unpaired dataset Tamper-Scene, we observe a 7.32\% increase in RecAcc compared to the state-of-the-art method STEEM~\cite{10719657}. For the synthetic dataset Tamper-Syn2k, since some of the ground truth images are not even discernible to the naked eye, our method fails to generate images that closely resemble the ground truth. This leads to a visual discrepancy and results in a higher FID.

In the second set of evaluations, we assess our method on common benchmarks for text recognition in terms of RecAcc metric, which is highly challenging for scene text editing. Using images in these datasets as the style images, we randomly sample a word that differs from the image annotation to serve as the target text for each style image. We then apply various models to perform scene text editing. Finally, we employ a more powerful text recognition model ABINet~\cite{fang2021read} to conduct text recognition on the edited images, as CRNN~\cite{shi2016end} shows limited performance on these more challenging datasets. Table~\ref{tab:real_editing} presents the experimental results. Note that the recognition performance of ABINet~\cite{fang2021read} on the original images, labeled as \textit{Base}, is used as baseline implying the upper bound of recognition performance. We observe that our \mymodel demonstrates an average performance improvement of 45.0\% over MOSTEL~\cite{qu2023exploring}, utilizing the same real training data MLT2017. It is encouraging that, when fine-tuned on Union14M-L dataset, our \mymodel achieves comparable performance with the upper bound on `SVT', `IC15' and `SVTP', which further demonstrates the effectiveness of our method.

To further illustrate the effectiveness of our method, we also provide qualitative comparisons. Accordingly, we present several visual qualitative examples in Figure ~\ref{fig:examples}. It is clear that our method significantly outperforms other methods in terms of editing effects. Further visualization results can be found in the supplementary material.

\begin{figure*}
\centering
    \includegraphics[width=0.95\textwidth]{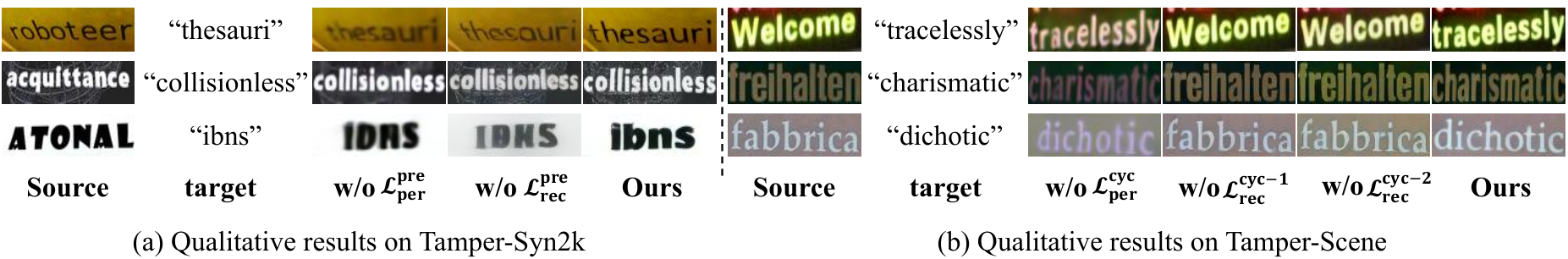}
\caption{Visualization examples of ablation study.}
\label{fig:ablations} 
\end{figure*}

\subsection{Ablation Study}
\noindent\textbf{Intrinsic Recognition.}
To validate the effectiveness of \mymodel's intrinsic recognition, which implicitly disentangles style and content while simultaneously ensuring content accuracy, we conduct a series of comparative experiments. The results are presented in Table~\ref{tab:ablation} and \ref{tab:ablation2}. 

First, to assess the disentanglement ability of our \mymodel, we conduct an experiment under the same conditions optimized for editing task, disregarding the model's recognition ability. Specifically, we achieve this by excluding the recognition loss during the training process. The comparison results are listed in Table~\ref{tab:ablation} \textit{w/o} $\mathcal{L}^{\text{pre}}_{\text{rec}}$ and \textit{Ours}. As can be easily seen, joint optimization of recognition and editing abilities during training can significantly enhance the editing performance of the model, particularly in terms of overall structure and realism, resulting in a 3.20 increase in SSIM and a 3.67 decrease in FID. %The comparison of the experimental outcomes, presented in Table~\ref{tab:ablation} \textit{w/o $\mathcal{L}_{\text{rec}}$} and \textit{Ours}, indicates that jointly optimizing the recognition ability with the editing ability during training substantially enhances the model's editing performance, especially for the overall structure and authenticity, as it yields an SSIM improvement of 3.20 and an FID improvement of 3.67. 

Second, to verify that the intrinsic recognition model offers superior content consistency compared to a conventional pre-trained external recognition model, we conduct an experiment incorporating the pre-trained model ABINet~\cite{fang2021read} solely to supervise the recognition of generated images. Results under identical training conditions are presented in Table~\ref{tab:ablation} \textit{w/ }$\mathcal{L}^{\text{per}}_{\text{orec}}$.  It can be observed that compared to the lack of supervised recognition task \textit{w/o} $\mathcal{L}^{\text{pre}}_{\text{rec}}$, using an external recognition model for supervision can improve model performance. However, the external recognition model can only constrain the consistency of the edited results in terms of content and does not achieve the decoupling between style and content like our intrinsic recognition model, thus resulting in slightly inferior performance.
%While the external recognition model does enhance the editing capabilities to some extent, our intrinsic recognition approach demonstrates superior performance due to its inherent ability to achieve disentanglement through intrinsic recognition.

\begin{table}
\caption{Ablation studies on $\mathcal{L}^{\text{pre}}$ applied in the fully-supervised pre-training stage using paired synthetic Tamper-Syn2k dataset. `w/o $\mathcal{L}^{\text{pre}}_{\text{mse}}$' denotes the absence of $\mathcal{L}^{\text{pre}}_{\text{mse}}$, while `w/ $\mathcal{L}^{\text{pre}}_{\text{orec}}$' means the utilization of pre-trained recognition supervisor.}
\label{tab:ablation}
\centering
\resizebox{0.65\linewidth}{!}{%
\renewcommand{\arraystretch}{1.2}
\begin{tabular}{l|cccc}
\toprule
\multirow{2}{*}{Methods} & \multicolumn{4}{c}{Tamper-Syn2k}  \\ \cmidrule{2-5} 
                         & MSE$\downarrow$    & PSNR$\uparrow$    & SSIM$\uparrow$   & FID$\downarrow$     \\ \midrule
                        w/o $\mathcal{L}^{\text{pre}}_{\text{per}}$ &  \textbf{0.0058}      &  \textbf{24.05}       &   \textbf{78.02}     &  70.06      \\
                        w/o $\mathcal{L}^{\text{pre}}_{\text{mse}}$ &  0.0086      &   21.94      &  73.79      &  \textbf{29.63}              \\
                        w/o $\mathcal{L}^{\text{pre}}_{\text{rec}}$& 0.0082       &    22.26     &   69.70     &   33.96   \\
                        w/ \enspace$\mathcal{L}^{\text{pre}}_{\text{orec}}$  &   0.0079     &  22.44       &   70.71     &  31.73      \\ \midrule
                         Ours&  0.0076      &  22.54       &  72.90      &  30.29 \\ \bottomrule
\end{tabular}%
}
\end{table}

\begin{table}
\centering
\caption{Ablation studies on $\mathcal{L}^{\text{cyc}}$ applied in the cyclic self-supervised fine-tuning stage using unpaired real dataset Tamper-Scene and paired real dataset ScenePair.}
\vspace{-4pt}
\label{tab:ablation2}
\resizebox{1\linewidth}{!}{%
\renewcommand{\arraystretch}{1.2}
\begin{tabular}{l|c|ccccc}
\toprule
\multirow{2}{*}{Methods}&Tamper-Scene &\multicolumn{5}{c}{ScenePair} \\ \cmidrule{2-7} 
                         & RecAcc$\uparrow$          & MSE$\downarrow$&PSNR$\uparrow$&SSIM$\uparrow$&FID$\downarrow$&RecAcc$\uparrow$          \\ \midrule
                        w/o $\mathcal{L}^{\text{cyc}}_{\text{per}}$ &  \textbf{96.23}             &0.0431&15.16&15.75&134.31&  88.52         \\
                        w/o $\mathcal{L}^{\text{cyc}}_{\text{mse}}$ &   83.79           & 0.0422         &14.79&17.64&97.00&    69.53  \\
                        w/o $\mathcal{L}^{\text{cyc-1}}_{\text{rec}}$ &   0.00      &  0.0500  &14.51&22.10&27.79&   4.38        \\
                        w/o $\mathcal{L}^{\text{cyc-2}}_{\text{rec}}$ &  0.00         &  0.0492   &14.76&22.10&24.61&      4.38     \\
                        w/o $\mathcal{L}^{\text{cyc}}$ &  69.01         &  0.0309   &16.66&36.81&46.60&      80.63    \\
                         \midrule
                         Ours&   86.12              & \textbf{0.0267}     &\textbf{17.35}&\textbf{46.09}&\textbf{41.37}&  \textbf{91.80}       \\ \bottomrule
\end{tabular}%
}

\end{table}

\noindent\textbf{Cyclic Self-Supervised Fine-tuning Strategy.}
\label{sec:cyclic_training}
Cyclic training enables \mymodel to optimize in a self-supervised manner on unpaired real-world data, thereby significantly enhancing its editing performance on real data.  %thereby significantly reducing the domain gap between the synthetic and real images after pre-training our method on synthetic data, as shown in Table \ref{tab:real_editing}. 
Specifically, as shown in Table~\ref{tab:real_editing} and \ref{tab:ablation2} \textit{w/o} $\mathcal{L}^{\text{cyc}}$, without fine-tuning on real datasets, \mymodel can only exhibit an average editing recognition accuracy of 55.7\% on commonly used text recognition benchmarks, 69.01\% on Tamper-Scene, and poor style consistency on ScenePair dataset. This is primarily due to the significant domain gap between the synthetic dataset used for pre-training and real-world scenarios. However, as shown in  Table~\ref{tab:real_editing}, when fine-tuned on the MLT2017 dataset using a cyclic training strategy, our approach achieves an average editing recognition accuracy of 81.8\%. Moreover, fine-tuning with the more complex and diverse Union14M-L dataset improves accuracy to 82.9\%, and significantly enhances style consistency on real-world dataset ScenePair, as shown in Table~\ref{tab:ablation2} \textit{Ours}. These results highlight the considerable potential for performance gains in our method.
\begin{table*}
\caption{Performance improvements of classical recognition models yielded from fine-tuning with edited bad cases from scene text editing models as data augmentation. All methods are pre-trained on Union14M-L.}
\label{tab:finetune}
\centering
\resizebox{0.95\linewidth}{!}{%
\begin{tabular}{l|l|lllllll|l}
\toprule
\multirow{2}{*}{Methods} &Augmentation & \multicolumn{7}{c|}{Union14M-Benchmark}    &\multirow{2}{*}{Avg.}                                   \\ \cmidrule{3-9} 
                       &Model  & Curve & Multi-Oriented & Artistic & Contextless & Salient & Multi-Words & \multicolumn{1}{l|}{General} &  \\ \midrule
& $-$ &   73.0   &    51.0    &   64.6   &  72.7    &   70.4   &    61.6    & \multicolumn{1}{l|}{77.9}    &  67.3   \\
ABINet~\cite{fang2021read}&  MOSTEL~\cite{qu2023exploring}                       &    73.7 \textcolor{green}{+0.7}  &   53.1 \textcolor{green}{+2.1}     &   65.0 \textcolor{green}{+0.4}  &  \textbf{73.8 \textcolor{green}{+1.1}}   &   72.2 \textcolor{green}{+1.8}  & 60.1 \textcolor{red}{-1.5}      & \multicolumn{1}{l|}{78.0 \textcolor{green}{+0.1}}    & 68.0 \textcolor{green}{+0.7}    \\
&     Ours                   &  \textbf{74.5 \textcolor{green}{+1.5}}   &   \textbf{54.5 \textcolor{green}{+3.5}}    &  \textbf{65.8 \textcolor{green}{+1.2}}   &  73.7 \textcolor{green}{+1.0}   & \textbf{73.9 \textcolor{green}{+3.5}}    &  \textbf{65.4 \textcolor{green}{+3.8}}     & \multicolumn{1}{l|}{\textbf{78.8 \textcolor{green}{+0.9}}}    &  \textbf{69.5 \textcolor{green}{+2.2} }  \\ \midrule
& $-$ &  81.4   &    71.4    &   72.0   &  82.0    &   78.5   &    82.4    & \multicolumn{1}{l|}{82.5}    &  78.6   \\
MAERec-S~\cite{Jiang2023RevisitingST}  &    MOSTEL~\cite{qu2023exploring}                   &    84.0 \textcolor{green}{+2.6}  &   72.0 \textcolor{green}{+0.6}     &   72.9 \textcolor{green}{+0.9}   &  80.2 \textcolor{red}{-1.8}    &   79.1 \textcolor{green}{+0.6}   & 82.3 \textcolor{red}{-0.1}      & \multicolumn{1}{l|}{82.1 \textcolor{red}{-0.4}}    &  78.9 \textcolor{green}{+0.3}   \\
 &      Ours                &  \textbf{85.0 \textcolor{green}{+4.6}}   &   \textbf{75.4  \textcolor{green}{+4.0}}   & \textbf{76.1 \textcolor{green}{+4.1}}   &  \textbf{82.9 \textcolor{green}{+0.9} }   & \textbf{80.9 \textcolor{green}{+2.4}}    &  \textbf{84.3 \textcolor{green}{+1.9} }    & \multicolumn{1}{l|}{\textbf{83.2 \textcolor{green}{+0.7}}}    &  \textbf{81.1 \textcolor{green}{+2.5}}   \\ \bottomrule
\end{tabular}%
}

\end{table*}

Additionally, during the cyclic training stage, our intrinsic recognition ensures content consistency between the target text and the edited image. As shown in Table~\ref{tab:ablation2} and Figure~\ref{fig:ablations}, when our model does not utilize $\mathcal{L}^{\text{cyc-1}}_{\text{rec}}$ and $\mathcal{L}^{\text{cyc-2}}_{\text{rec}}$ for constraints, it tends to learn an identical mapping from the original image to itself, which directly results in the loss of our model's ability to perform scene text editing. This demonstrates that, through the intrinsic recognition supervision in the cyclic training stage, our model is capable of decomposing content and style on real-world data, while ensuring content consistency.

\noindent\textbf{Loss Functions.}
In addition to recognition loss, MSE and Perception losses also play important roles for preserving style. We also conduct ablation studies to specifically discuss the effectiveness of different losses in the pre-training and cyclic training stages. The results are listed in Table~\ref{tab:ablation} and Table~\ref{tab:ablation2} respectively. %the role of other losses in different stages.

%Tables \ref{tab:ablation} and \ref{tab:ablation2} present ablation studies in the fully-supervised pre-training stage and the cyclic self-supervised fine-tuning stage, respectively, to verify the effectiveness of the loss functions $\mathcal{L}^\text{pre}$ and $\mathcal{L}^\text{cyc}$. %These analyses highlight each component's contribution to overall model performance.
As shown in Table~\ref{tab:ablation}, in the pre-training stage, \textit{w/o} $\mathcal{L}^{\text{pre}}_{\text{per}}$  or  \textit{w/o} $\mathcal{L}^{\text{pre}}_{\text{mse}}$  will significantly result in poorer editing performance.  Specifically, $\mathcal{L}^{\text{pre}}_{\text{per}}$ enhances the visual realism of generated images, as indicated by lower FID scores, while $\mathcal{L}^{\text{pre}}_{\text{mse}}$ ensures the pixel-level similarity indicated by MSE, PSNR, and SSIM. The same phenomenon can be seen in Figure~\ref{fig:ablations} (a).

%To investigate the effect of $\mathcal{L}^{\text{pre}}$, we evaluate the effects of removing perceptual loss $\mathcal{L}^{\text{pre}}_{\text{per}}$ and MSE loss $\mathcal{L}^{\text{pre}}_{\text{mse}}$ by observing there impacts on MSE, PSNR, SSIM, and FID scores on Tamper-Syn2k in Table~\ref{tab:ablation}. Specifically, $\mathcal{L}^{\text{pre}}_{\text{per}}$ enhances the visual realism of generated images, as indicated by lower FID scores, while $\mathcal{L}^{\text{pre}}_{\text{mse}}$ ensures the pixel-level similarity indicated by MSE, PSNR, and SSIM.
%Table~\ref{tab:ablation} %\textit{w/o}$\mathcal{L}^{\text{pre}}_{\text{per}}$ demonstrates that omitting perpetual loss leads to significantly higher MSE, PSNR, and SSIM scores but results in a poorer FID, reflecting reduced visual realism. Figure~\ref{fig:ablations} (a) \textit{w/o} $\mathcal{L}^{\text{pre}}_{\text{per}}$ further illustrates this degradation in visual quality, showing noticeably blurred images. Conversely, while MSE loss contributes to pixel-level fidelity, its removal improves perceived visual quality (lower FID) at the cost of worsened pixel-level metrics MSE, PSNR, and SSIM.

In the cyclic fine-tuning stage, cyclic perceptual loss ($\mathcal{L}^{\text{cyc}}_\text{per}$) and cyclic MSE loss ($\mathcal{L}^{\text{cyc}}_{\text{mse}}$) supervise the pixels of the reproduced results. Without these losses, as shown in Table~\ref{tab:ablation2} \textit{w/o} $\mathcal{L}^{\text{cyc}}_{\text{per}}$ and \textit{w/o} $\mathcal{L}^{\text{cyc}}_{\text{mse}}$, though \mymodel can achieve a better recognition accuracy on Tamper-Scene, it fails to preserve the style and tend to generate the targeted content in an extraneous style, as illustrated in Figure~\ref{fig:ablations} (b) \textit{w/o} $\mathcal{L}^{\text{cyc}}_{\text{per}}$.

\subsection{Targeted Data Augmentation for Recognition}
\label{sec:finetune_app}
In this section, we address the generation of targeted training data that simulates challenging cases encountered by the recognition model. This data augmentation strategy aims to fine-tune the recognition model, thereby increasing its accuracy and robustness in real-world applications. By addressing specific recognition errors, this targeted fine-tuning strategy markedly improves both general recognition models, such as ABINet~\cite{fang2021read}, and state-of-the-art recognition models, including MAERec-S~\cite{Jiang2023RevisitingST}. The results in Table~\ref{tab:finetune} show a significant improvement: the average recognition accuracy of ABINet~\cite{fang2021read} and MAERec-S~\cite{Jiang2023RevisitingST} increase by 2.2\% and 2.5\%, respectively, with our augmentation method. In contrast,  MOSTEL~\cite{qu2023exploring} only leads to an improvement of 0.7\% and 0.3\%, respectively. This result illustrates that our targeted data augmentation technique using our method significantly enhances the performance of recognition, even when the recognition model already achieves strong results. Implementation details will be provided in the supplementary.

% This enhancement underscores the effectiveness of tailored data in improving model robustness and accuracy in real-world applications.

% To evaluate the effectiveness of this approach, we use the Union14M-L dataset on two open-sourced recognition models ABINet~\cite{fang2021read} and MAERec-S~\cite{Jiang2023RevisitingST}, comparing our method with MOSTEL~\cite{qu2023exploring} to validate its superiority. For instance, on the ABINet~\cite{fang2021read} model, we first evaluate the pre-trained ABINet~\cite{fang2021read} on the Union14M-L dataset by testing on its evaluation set and identifying cases of incorrect recognition ("bad cases"). These bad cases are then modified using our method or MOSTEL~\cite{qu2023exploring}, generating additional text images that maintain a similar style but contain varied content for further fine-tuning of ABINet~\cite{fang2021read}. 

% In implementation, we employ each scene text editing model to randomly generate five variations per bad case, creating images with different textual content while retaining the original style. This process results in 450k and 300k augmented images for ABINet~\cite{fang2021read} and MAERec-S~\cite{Jiang2023RevisitingST}, respectively. The models are subsequently fine-tuned on a combination of these augmented datasets and the Union14M-L dataset.

\label{sec:pretrain_app}

% \section{Limitation and Conclusion}
% \noindent\textbf{Limitation.}
% A potential limitation of our method as well as most other methods for scene text editing lies in the limited performance when editing images with extremely large text curvature. This limitation is mainly attributed to the scarcity of such data in synthetic training data. Thus, constructing high-quality synthetic data that can simulate diverse challenging scenarios in real-world scenarios is essential, which we intend to explore in future work.

% \noindent\textbf{Conclusion.} In this work, we present \textbf{RS-STE} for scene text editing, which conducts scene text recognition and text editing in a unified framework, thereby eliminating the intricate model design for explicit disentanglement of background style and text content. Moreover, we devise the Cyclic Self-Supervised Fine-Tuning strategy, which is able to fine-tune our \mymodel using unpaired real-world data, significantly enhancing its generalizability to real-world scenarios. Extensive experiments validate the effectiveness of the proposed \mymodel.

\section{Conclusion}
In this work, we present \textbf{RS-STE}, which conducts recognition-synergistic scene text editing in a unified framework, thereby eliminating the intricate model design for explicit disentanglement of background style and text content. Moreover, we devise the Cyclic Self-Supervised Fine-Tuning strategy, which is able to fine-tune our \mymodel using unpaired real-world data, significantly enhancing its generalizability to real-world scenarios. Extensive experiments validate the effectiveness of the proposed \mymodel.

\section*{Acknowledgements}
This work was supported in part by the National Natural Science Foundation of China (62372133, 62125201, U24B20174), in part by Shenzhen Fundamental Research Program (Grant NO. JCYJ20220818102415032).
\clearpage
{
    \small
    \bibliographystyle{ieeenat_fullname}
    \bibliography{main}
}

\clearpage
\setcounter{page}{1}
\setcounter{section}{0} 
\setcounter{table}{5}
\setcounter{figure}{5}
\maketitlesupplementary
\appendix

\section{Summary}
This supplementary material comprises four components: (1) detailed descriptions of MMPD in our \mymodel; (2) implementation details of the fine-tuning stage of detokenizer and data augmentation for recognition; (3) additional ablation studies on model size and the feature representation approach; (4) limitation and analysis; and (5)  more visualization examples generated by various scene text editing methods and our \mymodel.
\section{Details of MMDP}
As described in Section~\ref{sec:method}, the input of MMPD can be denoted as $[\mathbf{E}^{\text{t}}_{T_B}, \mathbf{E}^{\text{i}}_{I_A}, \mathbf{E}^{\text{t}}_{\text{query}}, \mathbf{E}^{\text{i}}_{\text{query}}] \in \mathbb{R}^{2(L+N) \times C}$, where $L$ presents the length of the text embeddings and $N$ presents the length of the flattened image embeddings. In our configuration, we set $L=32$, $N=256$ and $C=384$.  Our MMPD consists of 12 transformer blocks, each of which includes a layer normalization layer, causal self-attention with 6 heads, and a fully connected layer.

\section{More Implementation Details}

\noindent\textbf{Fine-tuning Stage of Detokenizer.} We initialize the pre-trained VAE from LDM~\cite{rombach2022high} using configuration parameters $f=4$, $Z=8192$ and $d=3$. To improve the decoder's performance in reconstructing text images from continuous features, we fine-tune the VAE on our training datasets. Specifically, we remove the codebook-related components from the pre-trained model and train it for 100k iterations using the Adam~\cite{kingma2014adam} optimizer with a batch size of 256, and a learning rate of $1.25\times 10^{-3}$. The reconstruction performance of the VAE before and after fine-tuning on the evaluation dataset Tamper-Syn2k and ScenePair is shown in Tables~\ref{tab:vae_finetune} and ~\ref{tab:vae_finetune2}. Compared to the pre-trained VAE, the fine-tuned VAE demonstrates better image reconstruction performance for text images. This metric also indicates the upper limit of the image editing performance when using the VAE decoder as an Image Detokenizer.

\begin{figure*}[!htb]
\centering
    \includegraphics[width=0.8\textwidth]{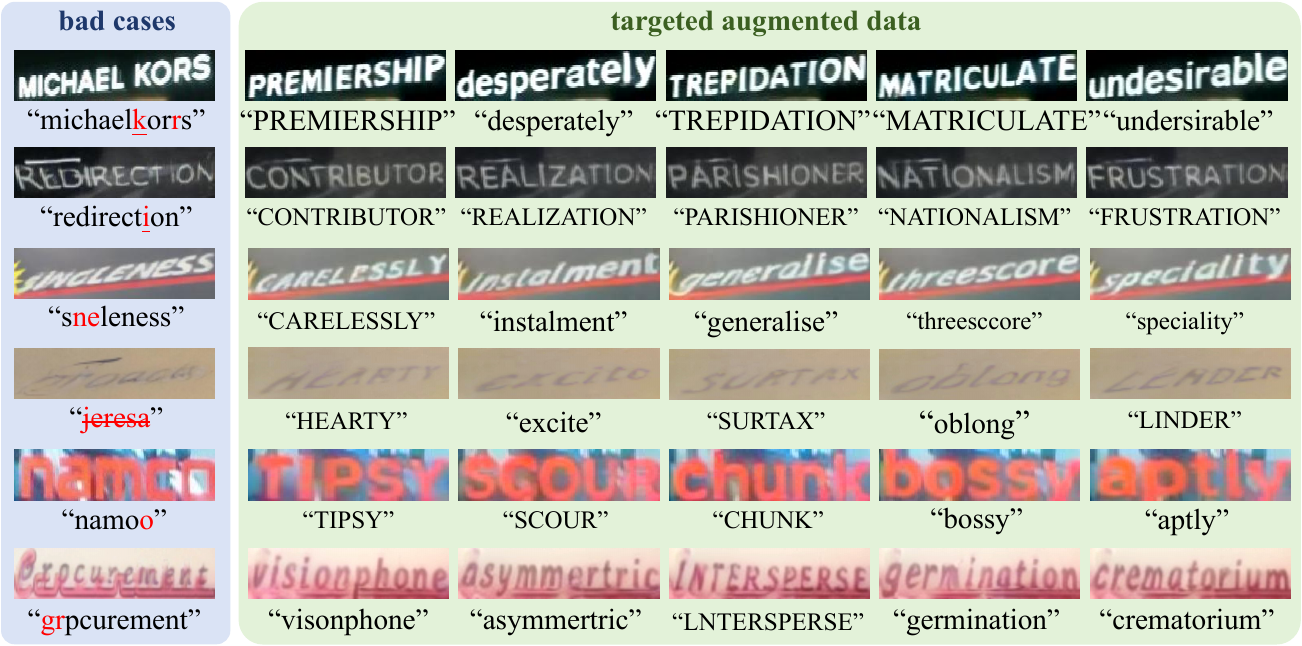}
\caption{The visualization of targeted augmented data generated by \mymodel from bad cases of recognition model ABINet~\cite{fang2021read}.}
\label{fig:augmented} 
\end{figure*}

\noindent\textbf{Details of Data Augmentation for Recognition.} To evaluate the effectiveness of our data augmentation strategy, we use the Union14M-L dataset on classical recognition model ABINet~\cite{fang2021read}, and state-of-the-art recognition model MAERec-S~\cite{Jiang2023RevisitingST}. We compare our method with MOSTEL~\cite{qu2023exploring} to validate its superiority. For instance, on the ABINet~\cite{fang2021read} model, we first evaluate the pre-trained ABINet~\cite{fang2021read} on the Union14M-L dataset by testing on its evaluation set and identifying cases of incorrect recognition ("bad cases"). These bad cases are then modified using our method or MOSTEL~\cite{qu2023exploring}, generating additional text images that maintain a similar style but contain varied content for further fine-tuning of ABINet~\cite{fang2021read}. We visualize some of the targeted augmented data generated by \mymodel in Figure~\ref{fig:augmented}.

In implementation, we employ each scene text editing model to randomly generate five variations per bad case, creating images with different textual content while retaining the original style. Subsequently, we utilize the corresponding pre-trained recognition models to recognize the generated targeted augmented images. Any data with an edit distance between the recognition result and the ground truth exceeding one-third of the word length is discarded. This process results in about 250k and 170k augmented images for ABINet~\cite{fang2021read} and MAERec-S~\cite{Jiang2023RevisitingST}, respectively. The models are subsequently fine-tuned on a combination of these augmented datasets and the Union14M-L dataset. 
% \usepackage{multirow}
\begin{table}[t]
\caption{The image reconstruction performance of VAE before and after fine-tuning on Tamper-Syn2k.}
\label{tab:vae_finetune}
\centering
\resizebox{0.75\linewidth}{!}{%
\begin{tabular}{c|cccc}
\toprule
\multirow{2}{*}{Fine-tune} & \multicolumn{4}{c}{Tamper-Syn2k}  \\ \cmidrule{2-5} 
                           & MSE$\downarrow$ & PSNR$\uparrow$ & SSIM$\uparrow$ & FID$\downarrow$\\ \midrule
                         \XSolidBrush  &   0.00453     & 25.22       &   83.17     &   30.91      \\
                         \Checkmark  &   \textbf{0.00049}     &    \textbf{34.01}    &    \textbf{98.57}    &   \textbf{13.34}       \\ \bottomrule
\end{tabular}%
}
\end{table}
\begin{table}
\caption{The image reconstruction performance of VAE before and after fine-tuning on ScenePair.}
\label{tab:vae_finetune2}
\centering
\resizebox{0.75\linewidth}{!}{%
\begin{tabular}{c|cccc}
\toprule
\multirow{2}{*}{Fine-tune}  & \multicolumn{4}{c}{ScenePair} \\ \cmidrule{2-5} 
                           & MSE$\downarrow$ & PSNR$\uparrow$ & SSIM$\uparrow$ & FID$\downarrow$ \\ \midrule
                         \XSolidBrush    &   0.00169   &    29.77    &   90.87     &   19.34    \\
                         \Checkmark     &   \textbf{0.00064}     &    \textbf{34.00}    &   \textbf{97.26}    &   \textbf{4.10}    \\ \bottomrule
\end{tabular}%
}
\end{table}

\begin{table*}
\caption{The impact of model scaling on editing performance of \mymodel. 'Tiny' denotes the 22.5M MMDP while 'Small' denotes the 85.5M one.}
\label{tab:supp_modelscale}
\centering
\resizebox{0.95\linewidth}{!}{%
\begin{tabular}{l|l|cccc|c|ccccc}
\toprule
\multirow{2}{*}{Model}& MMDP  & \multicolumn{4}{c|}{Tamper-Syn2k}   & Tamper-Scene& \multicolumn{5}{c}{ScenePair}\\ \cmidrule{3-12}
                         &\#Param.& MSE$\downarrow$ & PSNR$\uparrow$ & SSIM$\uparrow$ & FID$\downarrow$ & RecAcc$\uparrow$& MSE$\downarrow$& PSNR$\uparrow$& SSIM$\uparrow$& FID$\downarrow$&RecAcc $\uparrow$                  \\ \midrule
                     \mymodel-Tiny &22.5M& 0.0076  &  22.54    &  72.90    & \textbf{30.29}     &  86.12  &0.0267 &17.35&46.09&41.37   &\textbf{91.80}  \\
                    \mymodel-Small &85.5M&   \textbf{0.0072}  & \textbf{22.87} & \textbf{73.18}  & 31.34   & \textbf{94.14} &\textbf{0.0254} & \textbf{17.55}&\textbf{46.97} &\textbf{39.13} &91.56 \\ \bottomrule
\end{tabular}%
}
\end{table*}

% \usepackage{multirow}
\begin{table*}
\caption{The image reconstruction performance of continuous VAE and discrete VAE.}
\label{tab:vae_discrete}
\centering
\resizebox{0.65\linewidth}{!}{%
\begin{tabular}{l|cccc|cccc}
\toprule
\multirow{2}{*}{Condition} & \multicolumn{4}{c|}{Tamper-Syn2k} & \multicolumn{4}{c}{ScenePair} \\ \cmidrule{2-9} 
                           & MSE$\downarrow$ & PSNR$\uparrow$ & SSIM$\uparrow$ & FID$\downarrow$ & MSE$\downarrow$ & PSNR$\uparrow$ & SSIM$\uparrow$ & FID$\downarrow$ \\ \midrule
                         discrete  &   0.00146     & 30.18       &   93.79    &   21.35     &   0.00080   &    32.88    &   96.74     &   4.55    \\
                         continuous  &   \textbf{0.00049}     &    \textbf{34.01}    &    \textbf{98.57}    &   \textbf{13.34}     &  \textbf{0.00064}      &    \textbf{34.00}    &    \textbf{97.26}    &   \textbf{4.10}    \\ \bottomrule
\end{tabular}%
}
\end{table*}
\begin{table}
\caption{Text image editing performance with discrete and continuous feature representation methods.}
\label{tab:supp_discreteours}
\centering
\resizebox{0.7\linewidth}{!}{%
\renewcommand{\arraystretch}{1.2}
\begin{tabular}{l|cccc}
\toprule
\multirow{2}{*}{Methods} & \multicolumn{4}{c}{Tamper-Syn2k}  \\ \cmidrule{2-5} 
                         & MSE$\downarrow$    & PSNR$\uparrow$    & SSIM$\uparrow$   & FID$\downarrow$     \\ \midrule
                         discrete&    0.0167    &    19.03     &    70.57    &  46.73 \\
                         continuous&  \textbf{0.0076}      &  \textbf{22.54}       &  \textbf{72.90}      &  \textbf{30.29} \\ \bottomrule
\end{tabular}%
}
\end{table}
\begin{figure*}[!htb]
\centering
    \includegraphics[width=0.8\linewidth]{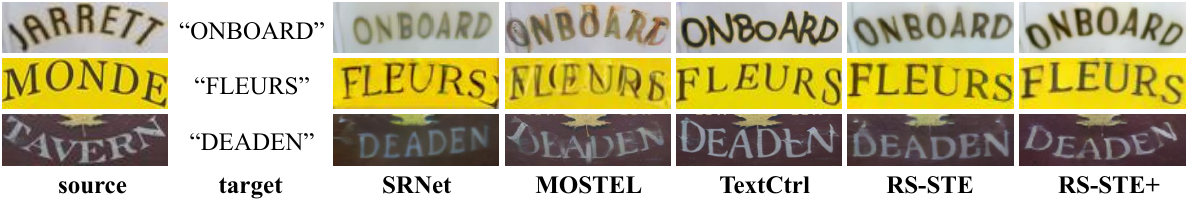}
\caption{Editing results of different methods on curved text.}
\label{fig:curve} 
\end{figure*}
\begin{figure}
\centering
    \includegraphics[width=0.9\linewidth]{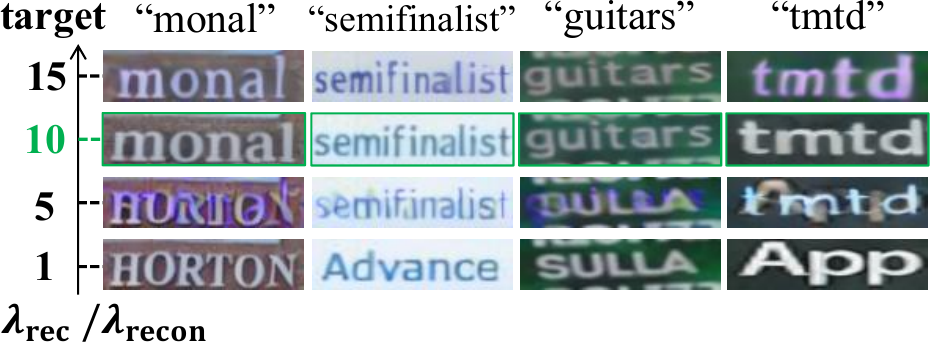}
\caption{Visualization examples of different ratio of recognition loss weight $\lambda_{\text{rec}}$ and reconstruction loss weight $\lambda_{\text{recon}}$.}
\label{fig:lossweight} 

\end{figure}
\section{More Ablation Studies}
\subsection{Effect of Model Size on Performance}
To further investigate the effect of model size on editing performance, we conduct experiments using an 85.5M MMDP model, configured with an embedding dimension of 768 and 12 attention heads. The results, presented in Table~\ref{tab:supp_modelscale}, demonstrate that increasing the model size significantly enhances the text editing capabilities of our approach. Therefore, in practical application, different model configurations can be selected based on a trade-off between computational resources and performance requirements.
\subsection{Discrete Feature Representation}
Since the pre-trained VAE from LDM~\cite{rombach2022high} utilizes Vector Quantization~\cite{van2017neural}, we also retain the fine-tuned VQ-VAE in our approach, using its encoder and codebook as the tokenizer and its decoder as the detokenizer. This design enables training on the discrete representations of both the source image and target text, leveraging the VAE's encoding and decoding mechanisms to their full potential. However, as illustrated in Table~\ref{tab:supp_discreteours}, our results indicate that the discrete feature encoding approach performs worse than the continuous encoding strategy adopted in our method.

This can primarily be attributed to two factors: (1) The discretization of images introduces information distortion, resulting in poorer reconstruction quality compared to continuous representations. As shown in Table~\ref{tab:vae_discrete}, for the given dataset, the reconstruction performance of the discrete form is inferior to that of the continuous form. (2) Continuous representations effectively mitigate the inherent decoding bias of the detokenizer. As discussed in Section~\ref{sec:method}, for continuous image features, reconstruction loss can be computed on the detokenized images, ensuring pixel-level accuracy in the final output. In contrast, for discrete representations, supervision can only be applied to the discretized image features decoded by the MMPD, leading to feature distortions during the detokenization process. 

\subsection{Loss Weights}
In the cyclic training stage described in Section~\ref{sec:training}, we observe that the ratio of the recognition loss weight, defined as $\lambda_{\text{rec}} = (\lambda_{6} + \lambda_{7}) / 2$, to the image reconstruction loss weight, defined as $\lambda_{\text{recon}} = (\lambda_{4} + \lambda_{5}) / 2$, plays a crucial role in ensuring content and style consistency. Consequently, we conduct an ablation study to examine the effects of varying this ratio, as shown in Figure~\ref{fig:lossweight}. Our findings indicate that a ratio close to 10 consistently produces high-quality images.

\section{Limitation and Analysis}
A potential limitation of our method as well as most other methods for scene text editing lies in the limited performance when editing images with extremely large text curvature, as shown in Figure~\ref{fig:curve}. This limitation is mainly attributed to the scarcity of such data in synthetic training data. To further investigate this issue, we train our model with additionally synthetic curved text samples generated using the synthesis engine mentioned in Section~\ref{sec:datasets}, and our method (\textbf{RS-STE+}) achieves robust curved text editing, which implies that such limitation arises from insufficient training data of curved text.

\begin{figure}[htpb]
\centering
    \includegraphics[width=0.85\linewidth]{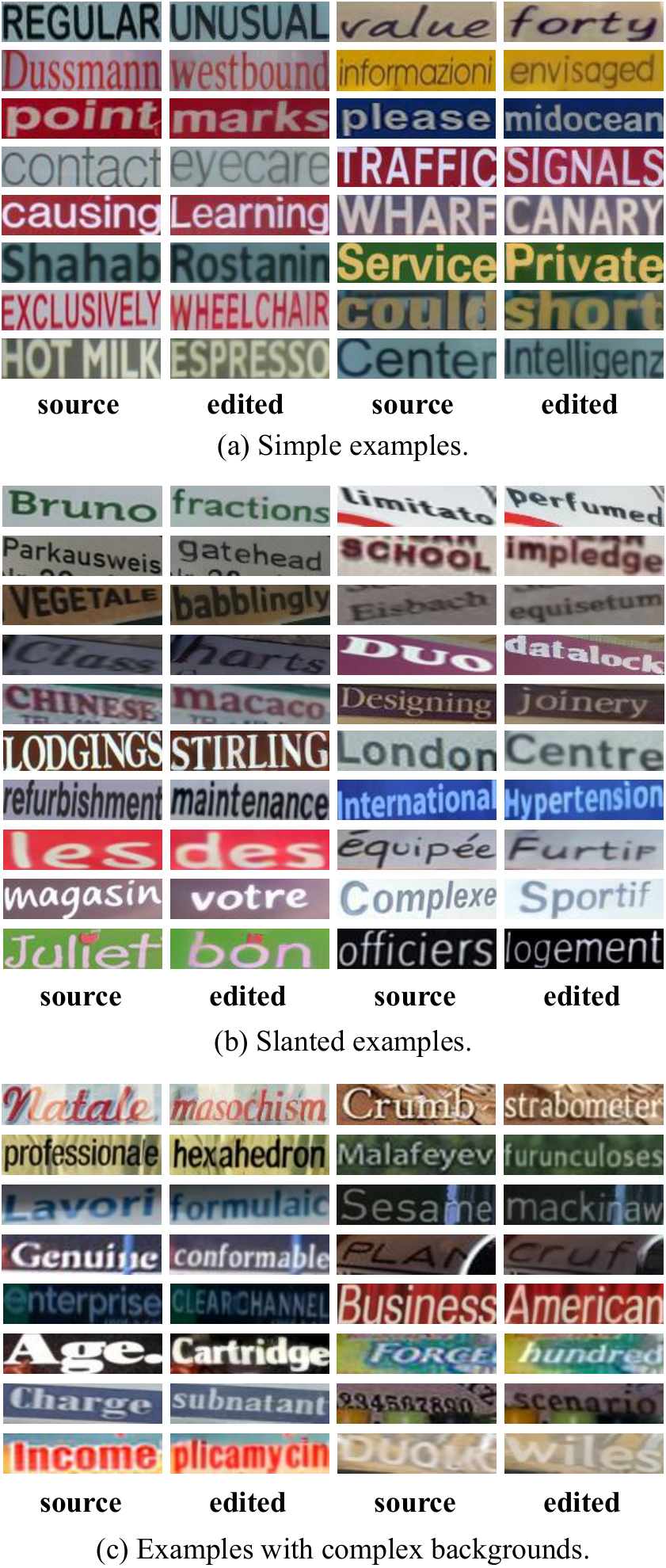}
\caption{More visualization examples edited by \mymodel on unpaired real-world dataset Tamper-Scene.}
\label{fig:supp_moreexamples} 
\end{figure}
\section{Visualization Examples of \mymodel}
To further demonstrate the superiority of our \mymodel, we include additional visualization results of the text images before and after editing with \mymodel, as illustrated in Figure~\ref{fig:supp_moreexamples}.
\label{sec:sup_visual}

\end{document}